# UNCERTAINTY IN QUANTUM RULE-BASED SYSTEMS

Vicente Moret-Bonillo [1,2], Isaac Fernández-Varela[1,2], Diego Álvarez-Estévez[3]

[1] Department of Computer Science. University of A Coruña.

[2] Research Center on Information and Communication Technology. Campus Elviña.

15071 A Coruña, Spain.

[3] Sleep Center & Clinical Neurophysiology at Haaglanden Medisch Centrum.

The Hague, Netherlands.

--..--

**Abstract.-** This article deals with the problem of the uncertainty in rule-based systems (RBS), but from the perspective of quantum computing (QC). In this work we first remember the characteristics of Quantum Rule-Based Systems (QRBS), a concept defined in a previous article by one of the authors of this paper, and we introduce the problem of quantum uncertainty. We assume that the subjective uncertainty that affects the facts of classical RBSs can be treated as a direct consequence of the probabilistic nature of quantum mechanics (QM), and we also assume that the uncertainty associated with a given hypothesis is a consequence of the propagation of the imprecision through the inferential circuits of RBSs. This article does not intend to contribute anything new to the QM field: it is a work of artificial intelligence (AI) that uses QC techniques to solve the problem of uncertainty in RBSs. Bearing the above arguments in mind a quantum model is proposed. This model has been applied to a problem already defined by one of the authors of this work in a previous publication and which is briefly described in this article. Then the model is generalized, and it is thoroughly evaluated. The results obtained show that QC is a valid, effective and efficient method to deal with the inherent uncertainty of RBSs.

--..--

## 1. Introduction

Uncertainty is one of the fundamental problems of artificial intelligence (AI). In particular, it is one of the essential problems of the so-called Rule-Based Systems (RBS), and at the same time one of the most complex to deal with.

Broadly speaking, we can consider that the origin of the uncertainty is related to one or several of the following causes [Lindley, 2014]:

- It may happen that the information available is incomplete. In many cases the information available is not sufficient to make a categorical decision.
- Sometimes the available information we handle is wrong. Not always the information we manage is completely true.

- The information we use is usually imprecise. In many domains there are data that are difficult to quantify.
- Normally the real world is non-deterministic. Intelligent systems are not always governed by deterministic laws, so that general laws are not always applicable. Many times the same causes produce different effects without there being any apparent explanation.
- Our model is often incomplete. There are many phenomena whose cause is unknown. In addition, the lack of agreement between experts in the same field is frequent. Both circumstances make it difficult to include knowledge in a RBS.
- It may happen that, even if our model is complete, it contains inaccurate information. Despite the large number of procedures that exist, any model that tries to quantify the uncertainty needs to include a large number of parameters. An example is the case of Bayesian networks [Pearl, 1986], in which we need to specify all a priori and conditional probabilities. However, a large part of this information is not usually available, so it must be estimated subjectively.

In summary, the treatment of uncertainty is, together with the representation of knowledge and learning, one of the fundamental problems of artificial intelligence (AI). In this article we will describe a quantum method to represent the uncertainty that may appear in the so-called Quantum Rule-Based Systems (QRBS) [Moret-Bonillo, 2018]. For this we will consider the following restrictions:

- We will consider that the rules of the knowledge base are written in a categorical way. For example: A and B $\rightarrow$ C
- In the resolution of a problem the facts may be affected by imprecision and, nevertheless, we have to use the rules of the knowledge base to obtain valid inferences. For example: we have the fact $\mathbb{A}$, which looks like A but it is not exactly A. We also have the fact $\mathbb{B}$, which looks like B but it is not exactly B. The question implies to be able of making inferences with

    o $\mathbb{A}$

    o $\mathbb{B}$

    o A and B $\rightarrow$ C

- Uncertainty arises as a consequence of the propagation of imprecision through the inferential network.

The organization of the material presented in this paper is as follows: First of all we remember that, based on the quantum circuit model [Marinescu, 2008], we can construct QRBSs that behave analogously to conventional RBSs. We also describe the quantum logic gates RQ-AND and RQ-OR, necessary for the construction of QRBSs, and we verify that the results of their application coincide, given the probabilistic nature of QM, almost exactly with the Truth Tables of the classic logical operators AND and OR.

Here we recall some of the fundamental properties of Quantum Computation [Nielsen, 2000]. These properties are essential in the design and construction of QRBSs.

The next section directly addresses the problem of the uncertainty in QRBSs. However, the followed approach has deficiencies that should be corrected. This correction is carried out in the following section, which concludes with the proposal of a universal model. After testing the behavior of the proposed model with several examples, the article presents the results obtained after the experimentation carried out, and concludes with the required discussion and with the bibliography used.

## 2. Classical Rule-Based Systems and Quantum Rule-Based Systems

In a recent article [Moret-Bonillo, 2018], Quantum Rule-Based Systems (QRBS) are defined as those Rule-Based Systems (RBS) that use the formalism of Quantum Mechanics (QM) for representing knowledge and for making inferences. Just to remember some key concepts, let us consider the following set of rules:

- R1: IF A and B THEN X
- R2: IF X or C THEN Y
- R3: IF Y and (D or E) THEN R

In conventional RBSs, any categorical rule can be represented by the logical operators {and, or, not} that relate statements that are always true. The truth tables of these conventional logical operators are the following (Table 1).

| X | Y | NOT X | X AND Y | X OR Y |
|---|---|-------|---------|--------|
| 0 | 0 | 1 | 0 | 0 |
| 0 | 1 | 1 | 0 | 1 |
| 1 | 0 | 0 | 0 | 1 |
| 1 | 1 | 0 | 1 | 1 |

*Table 1. Truth Values of the Classic Logical Operators NOT, AND and OR.*

Thus, rule R1 should be interpreted as follows:

- If statement A is true, and statement B is true, then we can conclude without uncertainty that statement X is true.

The three previously defined rules can be represented classically by means of the inferential circuit of figure 1.

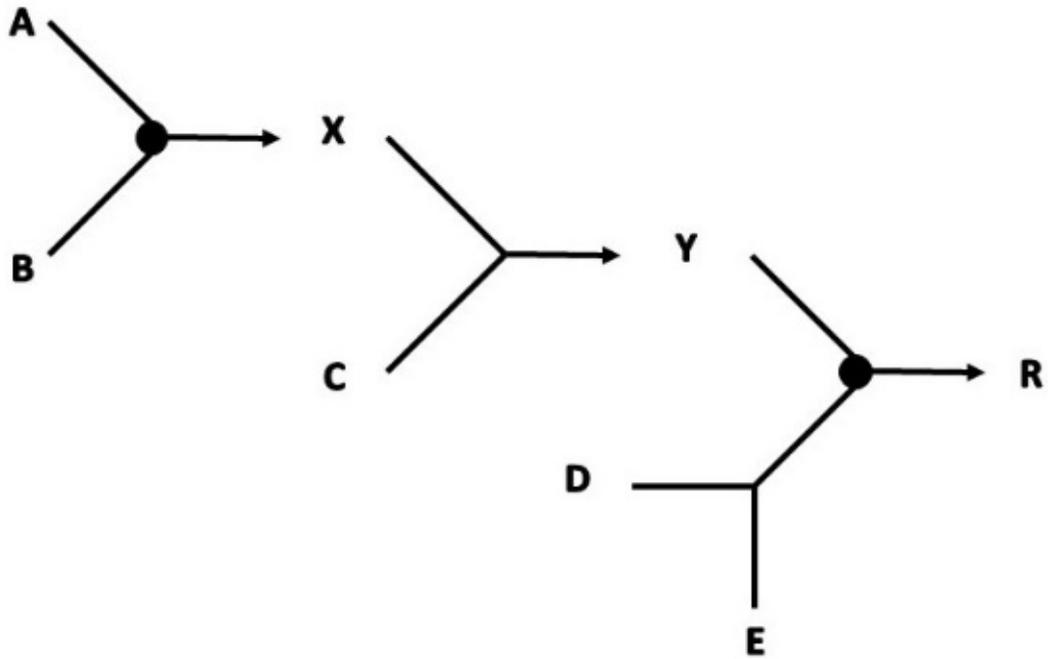

*Figure 1. A classic inferential circuit.*

However, if we choose the formalism of Quantum Computing we need reversible quantum gates to represent the previous inferential circuit [Moret-Bonillo, 2017]. More specifically, and in this case, according to the model of quantum circuits of IBM [IBM, 2018] the quantum representation of the circuit of Figure 1 would be that of Figure 2.

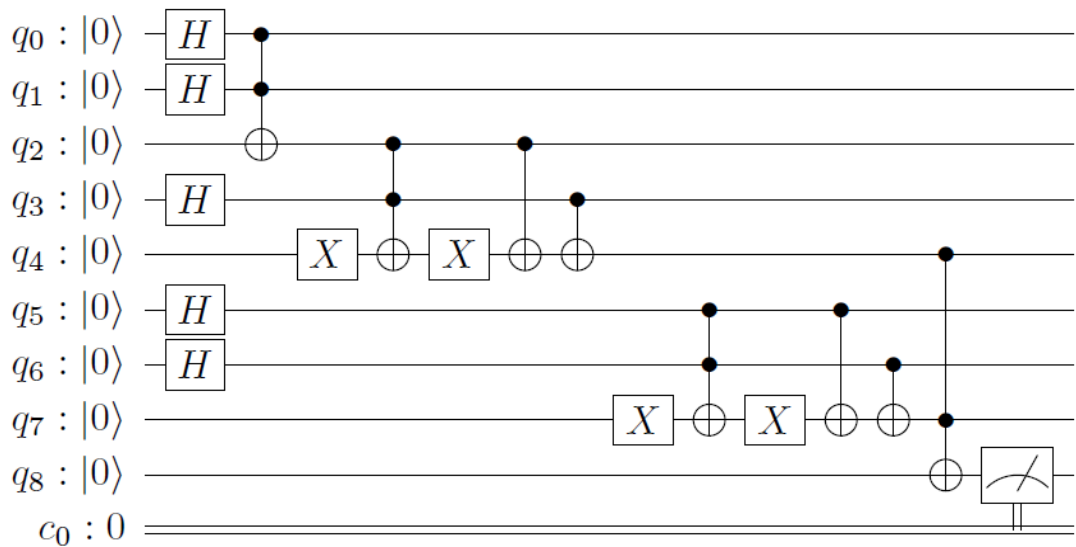

*Figure 2. The quantum inferential circuit equivalent to that of figure 1.*

In the quantum inferential circuit of Figure 2 we identify the following elements:

- Hadamard gates, used to achieve superpositions of states.
- CCN gates, which perform double-controlled NOT operations.
- X gates, which perform a conventional NOT on the state of the corresponding line.
- CN gates, which perform controlled NOT operations.
- An element of measurement that inform us about the result of the operations of the whole circuit.

Figure 3 illustrates all the mentioned elements related to the architecture shown in Figure 2. The reader can find a detailed description of all these elements and the corresponding gates in [Moret-Bonillo, 2017].

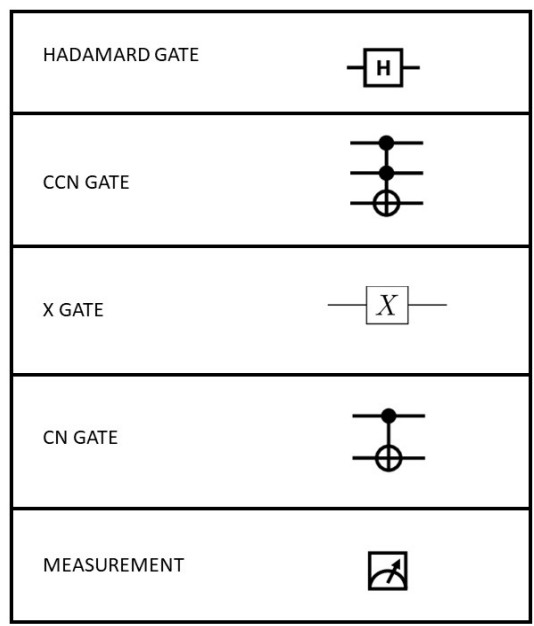

*Figure 3. Elements and gates used in the architecture of figure 2.*

The combination of quantum gates allows us to design and implement quantum logical operators. In this regard, Figure 4 shows the architecture of a quantum gate {and} -to which we have called RQ-AND-, and Table 2 the corresponding results obtained after 8192 executions in the quantum simulator of IBM. Similarly, Figure 5 shows the architecture of a quantum gate {or} -to which we have called RQ-OR-, and Table 3 the corresponding results obtained after 8192 executions in the IBM Quantum Simulator.

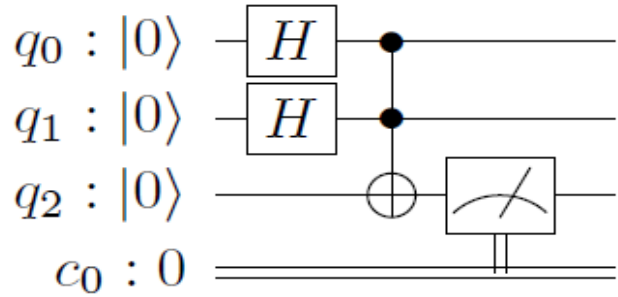

*Figure 4. Architecture of a quantum gate RQ-AND*

| Input Vector | Input Truth Table | Output Truth Table | Measured Percentage | Estimated Percentage | Precision |
|---|---|---|---|---|---|
| 000 | 00 | 0 | 25.0 | 25.0 | 1.000 |
| 010 | 01 | 0 | 24.8 | 25.0 | 0.992 |
| 100 | 10 | 0 | 24.9 | 25.0 | 0.996 |
| 111 | 11 | 1 | 25.3 | 25.0 | 0.988 |

*Table 2. Results obtained for the RQ-AND gate after 8192 executions in the IBM quantum simulator.*

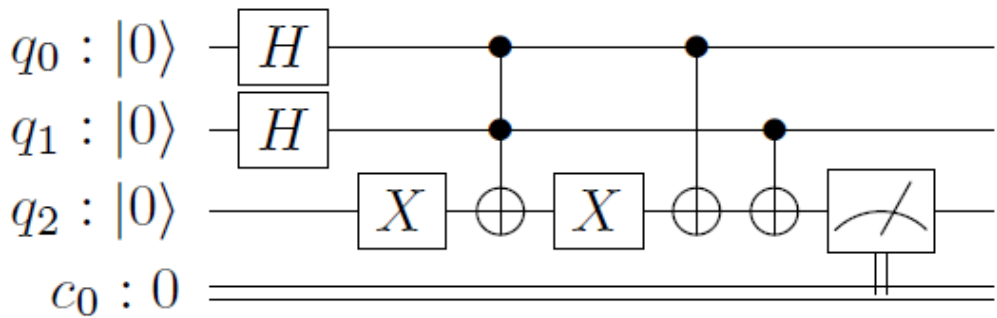

*Figure 5. Architecture of a quantum gate RQ-OR*

| Input Vector | Input Truth Table | Output Truth Table | Measured Percentage | Estimated Percentage | Precision |
|---|---|---|---|---|---|
| 000 | 00 | 0 | 25.0 | 25.0 | 1.000 |
| 011 | 01 | 1 | 24.8 | 25.0 | 0.992 |
| 101 | 10 | 1 | 24.9 | 25.0 | 0.996 |
| 111 | 11 | 1 | 25.3 | 25.0 | 0.988 |

*Table 3. Results obtained for the RQ-OR gate after 8192 executions in the IBM quantum simulator*

The detailed explanation of the operation of the quantum circuit of Figure 2 can be found in [Moret-Bonillo, 2018].

One issue that should be remembered is the non-deterministic nature of quantum mechanics, which means that quantum circuits produce probabilistic results. This circumstance makes us think about the problem of uncertainty, which is an inherent characteristic of Artificial Intelligence.

## 3. Basic Concepts of Quantum Computing

In Quantum Computing the information unit is the Qubit. The QuBit is a vector of a Hilbert space represented, in the Dirac notation, by a column matrix in such a way that a Bit 0 corresponds to a Ket 0, and a Bit 1 corresponds to a Ket 1 [Yanofsky, 2008]. The correspondence is the one shown below:

- $Bit\ 0 = \{0\} \rightarrow Ket\ 0 = |0\rangle = \begin{pmatrix} 1 \\ 0 \end{pmatrix}$

- $Bit\ 1 = \{1\} \rightarrow Ket\ 1 = |1\rangle = \begin{pmatrix} 0 \\ 1 \end{pmatrix}$

However, this correspondence is not complete, since quantum systems can be in what is called "coherent superposition", which means that a quantum system can be "simultaneously" in the states Ket 0 and Ket 1. To describe this peculiarity we need a State Function, $\psi$, that verifies the following restrictions:

- $|\psi\rangle = \begin{pmatrix} \alpha \\ \beta \end{pmatrix} = \alpha\ |0\rangle + \beta\ |1\rangle$

- $\alpha, \beta \in C\ \ being\ C\ the\ field\ of\ Complex\ Numbers$

- $|\alpha|^2 + |\beta|^2 = 1\ since\ the\ Total\ Probability\ has\ to\ be\ 1$

Thus, the QuBit:

- $|\psi\rangle = \frac{i}{\sqrt{2}}\ |0\rangle + \frac{1}{\sqrt{2}}\ |1\rangle = \frac{1}{\sqrt{2}} \begin{pmatrix} i \\ 1 \end{pmatrix}$

is a well defined quantum state, since:

- $\frac{i}{\sqrt{2}} \in C$
- $\frac{1}{\sqrt{2}} \in C$
- $\left|\frac{i}{\sqrt{2}}\right|^2 + \left|\frac{1}{\sqrt{2}}\right|^2 = 1$

In this QuBit the complex numbers:

- $\alpha = \frac{i}{\sqrt{2}}$
- $\beta = \frac{1}{\sqrt{2}}$

are called amplitudes of the state function. A fundamental aspect, which is a direct consequence of Heisenberg's indeterminacy principle, is that when we observe (or measure) a given QuBit in superposition, said QuBit loses its quantum properties and collapses, irreversibly, in what we have called classical bits. However, it does so with a certain probability. For example, if we measure the previous QuBit, we will have a probability of obtaining:

- $Bit\ \{0\}\ \rightarrow Prob\ \{0\} = \left|\frac{i}{\sqrt{2}}\right|^2 = 0.5$

- $Bit\ \{1\}\ \rightarrow Prob\ \{1\} = \left|\frac{1}{\sqrt{2}}\right|^2 = 0.5$

This property of quantum systems will have a decisive importance in our approach to the quantum uncertainty of Rule-based Quantum Systems.

## 4. Representing Uncertainty in Quantum Rule-Based Systems

In quantum mechanics, the Bloch sphere is a geometric representation of the pure state space of a two-level quantum system. By extension, the set of pure states of an arbitrary finite number of levels is also usually called the Bloch sphere. In this case the Bloch sphere is no longer a sphere, but it has a geometric structure known as a symmetric space. Consider Figure 6, which represents a Bloch Sphere.

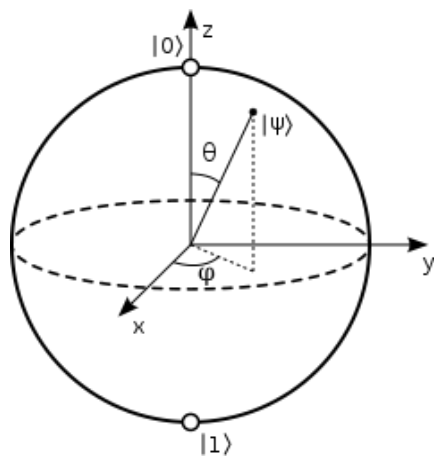

*Figure 6. Schematic representation of the Bloch Sphere.*

This sphere is important because it allows us to visualize the state of a QuBit. Note that it is a three-dimensional space in which the state of a QuBit $|\psi\rangle$ is represented by:

- The module, which is always 1. The meaning of this is that the QuBit is normalized.
- The angle $\theta$, which represents the displacement of the QuBit along the Z axis, from the north pole to the south pole of the sphere.
- The angle $\varphi$, which represents the phase of the system that is inherent in complex vector spaces.

Given that the Bloch sphere locates Ket 0 at the north pole, and since the direction of rotation on the Z axis is from top to bottom, we assume that a TRUE statement is represented by a Ket 0, and that a FALSE statement is represented by means of a Ket 1.

The question we are now asking is the following: assumed a given Quantum Rule-Based System, is there any quantum property that allows to represent the imprecision associated with the facts and the uncertainty associated with the rules?

To try to answer the previous question let's look at the angle $\theta$ of the Bloch sphere.

- When $\theta$ = 0 radians $\rightarrow |0\rangle \rightarrow$ The associated statement is true.
- When $\theta = \pi$ radians $\rightarrow |1\rangle \rightarrow$ The associated declaration is false.
- When $0 < \theta < \pi \rightarrow$ We are in a situation of coherent superposition, and the associated statement is neither true nor false, or - in an equivalent way - it is true and it is false simultaneously.

For practical reasons we will call "Credibility" to our confidence in a given fact. In such way that:

- Credibility = 100 $\rightarrow$ The fact is true

- Credibility = 0 $\rightarrow$ The fact is false

We also relate the concept of credibility with the concept of "Degree of Disbelief" associated to a given fact. The relation between these two concepts is as follows:

- Credibility = 100 $-$ Disbelief

For the reasons just explained, the quantification of Z displacements in a Bloch sphere could be used to quantify our credibility associated to a given fact.

In this context, the quantum approach requires the use of quantum gates and, following the indications of IBM Quantum Experience [IBM, 2018], we need the following gates:

- Gate H or Gate of Hadamard. This gate has the property that maps X $\rightarrow$ Z and Z $\rightarrow$ X and it is necessary to make superpositions. Its matrix is:

$$H = \frac{1}{\sqrt{2}}\begin{pmatrix} +1 & +1 \\ +1 & -1 \end{pmatrix}$$

- Gate S. This gate, which maps X $\rightarrow$ Y and Z $\rightarrow$ Z, is necessary to make complex superpositions, and its matrix is:

$$S = \begin{pmatrix} 1 & 0 \\ 0 & i \end{pmatrix}$$

- Gate T. This gate is a rotation $\pi$ / 4 around the Z axis, and is necessary for the universal control of the quantum system. Its matrix is:

$$T = \begin{pmatrix} 1 & 0 \\ 0 & \dfrac{1+i}{\sqrt{2}} \end{pmatrix}$$

- Gate Z. This gate, which is a rotation π around the Z axis, has the property that maps X → -X, and Z → Z , and generates a phase change. Its matrix is:

$$Z = \begin{pmatrix} +1 & 0 \\ 0 & -1 \end{pmatrix}$$

The product of any of these matrices with its corresponding conjugate transpose matrix is always the unit matrix, as shown below:

- $H \times H^\dagger = \frac{1}{\sqrt{2}}\begin{pmatrix} +1 & +1 \\ +1 & -1 \end{pmatrix} \times \frac{1}{\sqrt{2}}\begin{pmatrix} +1 & +1 \\ +1 & -1 \end{pmatrix} = \begin{pmatrix} 1 & 0 \\ 0 & 1 \end{pmatrix}$

- $S \times S^\dagger = \begin{pmatrix} 1 & 0 \\ 0 & i \end{pmatrix} \times \begin{pmatrix} +1 & 0 \\ 0 & -i \end{pmatrix} = \begin{pmatrix} 1 & 0 \\ 0 & 1 \end{pmatrix}$

- $T \times T^\dagger = \begin{pmatrix} 1 & 0 \\ 0 & \frac{1+i}{\sqrt{2}} \end{pmatrix} \times \begin{pmatrix} 1 & 0 \\ 0 & \frac{1-i}{\sqrt{2}} \end{pmatrix} = \begin{pmatrix} 1 & 0 \\ 0 & 1 \end{pmatrix}$

- $Z \times Z^\dagger = \begin{pmatrix} +1 & 0 \\ 0 & -1 \end{pmatrix} \times \begin{pmatrix} +1 & 0 \\ 0 & -1 \end{pmatrix} = \begin{pmatrix} 1 & 0 \\ 0 & 1 \end{pmatrix}$

But what is really important in this approach is that, as demonstrated in the IBM Quantum Experience, the product of certain doors involves the modification of the angle θ on the Z axis of the Bloch sphere, and also - after the process of measure- with a quantified probability. In this way it can be verified that:

- H x H = 0 radians
  - Rotation of 0 radians on the Z axis
  - Prob (0) = 1.00
  - Prob (1) = 0.00

- H x T x H = π/4 radians
  - Rotation of π/4 radians on the Z axis
  - Prob (0) = 0.85
  - Prob (1) = 0.15

- H x S x H = π/2 radians
  - Rotation of π/2 radians on the Z axis
  - Prob (0) = 0.50
  - Prob (1) = 0.50

- H x S x T x H = 3π/4 radians
  - Rotation of 3π/4 radians on the Z axis
  - Prob (0) = 0.15
  - Prob (1) = 0.85

- H x Z x H = π radians
  - Rotation of π radians on the Z axis
  - Prob (0) = 0.00
  - Prob (1) = 1.00

The previous products generate the following matrices that, of course, are trivially Hermitic:

$$H \times H = \begin{pmatrix} 1 & 0 \\ 0 & 1 \end{pmatrix}$$

$$H \times H \rightarrow \begin{pmatrix} 1 & 0 \\ 0 & 1 \end{pmatrix} \times \begin{pmatrix} 1 & 0 \\ 0 & 1 \end{pmatrix} = \begin{pmatrix} 1 & 0 \\ 0 & 1 \end{pmatrix}$$

$$H \times S \times H = \begin{pmatrix} \dfrac{1+\mathrm{i}}{2} & \dfrac{1-\mathrm{i}}{2} \\ \dfrac{1-\mathrm{i}}{2} & \dfrac{1+\mathrm{i}}{2} \end{pmatrix}$$

$$H \times S \times H \rightarrow \begin{pmatrix} \dfrac{1+\mathrm{i}}{2} & \dfrac{1-\mathrm{i}}{2} \\ \dfrac{1-\mathrm{i}}{2} & \dfrac{1+\mathrm{i}}{2} \end{pmatrix} \times \begin{pmatrix} \dfrac{1-\mathrm{i}}{2} & \dfrac{1+\mathrm{i}}{2} \\ \dfrac{1+\mathrm{i}}{2} & \dfrac{1-\mathrm{i}}{2} \end{pmatrix} = \begin{pmatrix} 1 & 0 \\ 0 & 1 \end{pmatrix}$$

$$H \times T \times H = \begin{pmatrix} \dfrac{2+\sqrt{2}+\mathrm{i}\sqrt{2}}{4} & \dfrac{2-\sqrt{2}-\mathrm{i}\sqrt{2}}{4} \\ \dfrac{2-\sqrt{2}-\mathrm{i}\sqrt{2}}{4} & \dfrac{2+\sqrt{2}+\mathrm{i}\sqrt{2}}{4} \end{pmatrix}$$

$$H \times T \times H \rightarrow \begin{pmatrix} \dfrac{2+\sqrt{2}+\mathrm{i}\sqrt{2}}{4} & \dfrac{2-\sqrt{2}-\mathrm{i}\sqrt{2}}{4} \\ \dfrac{2-\sqrt{2}-\mathrm{i}\sqrt{2}}{4} & \dfrac{2+\sqrt{2}+\mathrm{i}\sqrt{2}}{4} \end{pmatrix} \times \begin{pmatrix} \dfrac{2+\sqrt{2}-\mathrm{i}\sqrt{2}}{4} & \dfrac{2-\sqrt{2}+\mathrm{i}\sqrt{2}}{4} \\ \dfrac{2-\sqrt{2}+\mathrm{i}\sqrt{2}}{4} & \dfrac{2+\sqrt{2}-\mathrm{i}\sqrt{2}}{4} \end{pmatrix}$$
$$= \begin{pmatrix} 1 & 0 \\ 0 & 1 \end{pmatrix}$$

$$H \times S \times T \times H = \begin{pmatrix} \dfrac{2-\sqrt{2}+\mathrm{i}\sqrt{2}}{4} & \dfrac{2+\sqrt{2}-\mathrm{i}\sqrt{2}}{4} \\ \dfrac{2+\sqrt{2}-\mathrm{i}\sqrt{2}}{4} & \dfrac{2-\sqrt{2}+\mathrm{i}\sqrt{2}}{4} \end{pmatrix}$$

$$H \times S \times T \times H \rightarrow \begin{pmatrix} \dfrac{2-\sqrt{2}+i\sqrt{2}}{4} & \dfrac{2+\sqrt{2}-i\sqrt{2}}{4} \\ \dfrac{2+\sqrt{2}-i\sqrt{2}}{4} & \dfrac{2-\sqrt{2}+i\sqrt{2}}{4} \end{pmatrix} \times \begin{pmatrix} \dfrac{2-\sqrt{2}-i\sqrt{2}}{4} & \dfrac{2+\sqrt{2}+i\sqrt{2}}{4} \\ \dfrac{2+\sqrt{2}+i\sqrt{2}}{4} & \dfrac{2-\sqrt{2}-i\sqrt{2}}{4} \end{pmatrix}$$
$$= \begin{pmatrix} 1 & 0 \\ 0 & 1 \end{pmatrix}$$

$$H \times Z \times H = \begin{pmatrix} 0 & 1 \\ 1 & 0 \end{pmatrix}$$

$$H \times Z \times H \rightarrow \begin{pmatrix} 0 & 1 \\ 1 & 0 \end{pmatrix} \times \begin{pmatrix} 0 & 1 \\ 1 & 0 \end{pmatrix} = \begin{pmatrix} 1 & 0 \\ 0 & 1 \end{pmatrix}$$

We will analyze what has been done up to now with the help of the following example: let's apply H x T x H over $|0\rangle$. The result we get is:

$$\begin{pmatrix} \dfrac{2+\sqrt{2}+i\sqrt{2}}{4} & \dfrac{2-\sqrt{2}-i\sqrt{2}}{4} \\ \dfrac{2-\sqrt{2}-i\sqrt{2}}{4} & \dfrac{2+\sqrt{2}+i\sqrt{2}}{4} \end{pmatrix}\begin{pmatrix} 1 \\ 0 \end{pmatrix} = \left(\dfrac{2+\sqrt{2}+i\sqrt{2}}{4}\right)|0\rangle + \left(\dfrac{2-\sqrt{2}-i\sqrt{2}}{4}\right)|1\rangle$$

In this context, perhaps the most aggressive hypothesis we propose is the following: HYPOTHESIS: Given that in artificial intelligence all the information is real, we can simplify the previous expression if we consider only the modules of the amplitudes with the restriction that the total probability has to be 1.

Then we check that the restrictions are met:

- Modulus of $\left(\dfrac{2+\sqrt{2}+i\sqrt{2}}{4}\right)$=0.924
- Modulus of $\left(\dfrac{2-\sqrt{2}-i\sqrt{2}}{4}\right)$=0.383

Obviously: Total Probability = $0.383^2 + 0.924^2$ = 1.00

If we now do the same with all the combinations of gates described above, applying them on Ket 0, we obtain the following results:

- $(H \times H)\,|0\rangle = |0\rangle$
    - Probability (0) = 1.000
    - Probability (1) = 0.000
    - Total Probability = 1.00

- $(H \times T \times H)\,|0\rangle = 0.924\,|0\rangle + 0.383\,|1\rangle$
    - Probability (0) = 0.854
    - Probability (1) = 0.147
    - Total Probability = 1.00

- $(H \times S \times H)\,|0\rangle = 0.707\,|0\rangle + 0.707\,|1\rangle$
    - Probability (0) = 0.500
    - Probability (1) = 0.500
    - Total Probability = 1.00

- $(H \times S \times T \times H) \, |0\rangle = 0.383 \, |0\rangle + 0.924 \, |1\rangle$
  - Probability (0) = 0.147
  - Probability (1) = 0.854
  - Total Probability = 1.00

- $(H \times Z \times H) \, |0\rangle = |1\rangle$
  - Probability (0) = 0.000
  - Probability (1) = 1.000
  - Total Probability = 1.00

Apparently the approach works perfectly, or at least as expected, but it has several limitations. First, it allows only a discrete number of degrees of uncertainty. Secondly, to represent each degree of uncertainty we need, in a non-trivial way, to design and assemble a set of quantum gates. Can we do better?

## 5. Redefining Quantum Uncertainty

To define a general procedure capable of representing any degree of uncertainty (or certainty) it would be convenient to have a single quantum gate that, of course respecting all the restrictions imposed by quantum mechanics, brings us closer to the world of analog. In this context there are already several universal gates, but none of them explicitly works with imprecise information in the domain of artificial intelligence. In this regard, and taking into account what has been described so far, our proposal is as follows:

Let DELTA ($\delta$) be the degree of subjective disbelief that we can associate with a fact in a rule-based system. It is trivial that the parameter $\delta$ can be converted into an ALPHA angle ($\alpha$) that satisfies the restrictions of Z displacements. Now suppose that our subjective disbelief $\delta$ is defined in the closed interval [0, 100]. Obviously:

- If $\delta = 0 \rightarrow$ Our credibility in the fact is total $\rightarrow$ The fact is true
- If $\delta = 100 \rightarrow$ Our credibility in the negation of the fact is total $\rightarrow$ The fact is false
- If $0 < \delta < 100 \rightarrow$ There is subjective disbelief in the fact under consideration

The following equation establishes the correspondence between $\delta$ and $\alpha$, so that $\delta$ is compatible with the concept of subjective disbelief, and $\alpha$ is compatible with the restrictions imposed by the Bloch sphere:

- $\alpha = \frac{\pi \times \delta}{100} \, radians$

Now let us define THETA ($\theta$) = $(\pi - \alpha)$ / 2 as the angle of rotation, or displacement, in Z.

Table 4 illustrates the values of ALPHA ($\alpha$) - in degrees and in radians - as a function of the values of DELTA ($\delta$) -defined in the interval [0, 100], and the corresponding values of THETA ($\theta$) - expressed in radians-

| DELTA<br>(Subjective Disbelief) | ALPHA<br>(Degrees) | ALPHA<br>(Radians) | THETA<br>(Radians) |
|---|---|---|---|
| 0 | 0 | 0 | $\pi/2$ |
| 25 | 45 | $\pi/4$ | $3\pi/8$ |
| 50 | 90 | $\pi/2$ | $\pi/4$ |
| 75 | 135 | $3\pi/4$ | $\pi/8$ |
| 100 | 180 | $\pi$ | 0 |

*Table 4. Correspondence between the values of the parameters DELTA, ALPHA and THETA.*

We will now define, based on the angle $\theta$, the following Matrix:

- $M(\theta) = \begin{pmatrix} \sin(\theta) & \cos(\theta) \\ \cos(\theta) & -\sin(\theta) \end{pmatrix}$

This matrix verifies that:

$$\begin{pmatrix} \sin\theta & \cos\theta \\ \cos\theta & -\sin\theta \end{pmatrix} \times \begin{pmatrix} \sin\theta & \cos\theta \\ \cos\theta & -\sin\theta \end{pmatrix} = \begin{pmatrix} 1 & 0 \\ 0 & 1 \end{pmatrix}$$

In this context:

- If $\alpha = 0 \rightarrow M(\theta) |0\rangle = M(\pi/2) |0\rangle = |0\rangle$
- If $\alpha = \pi/4 \rightarrow M(\theta) |0\rangle = M(3\pi/8) |0\rangle = 0.924 |0\rangle + 0.383 |1\rangle$
- If $\alpha = \pi/2 \rightarrow M(\theta) |0\rangle = M(\pi/4) |0\rangle = 0.707 |0\rangle + 0.707 |1\rangle$
- If $\alpha = 3\pi/4 \rightarrow M(\theta) |0\rangle = M(\pi/8) |0\rangle = 0.383 |0\rangle + 0.924 |1\rangle$
- If $\alpha = \pi \rightarrow M(\theta) |0\rangle = M(0) |0\rangle = |1\rangle$

These results coincide exactly with those obtained from the quantum gate combination method, but -in this case- we have only needed a single matrix. Obviously:

- If $\alpha = 0 \rightarrow$ there is no disbelief in the fact, the credibility is total and the fact is true
- If $\alpha = \pi \rightarrow$ there is no disbelief in the negation of the fact, the credibility is none and the fact is false

## 6. Consistency of the approach

The proposed approach has been applied on a set of examples to check the consistency of the model. Table 5 illustrates the results obtained for different ALPHA values.

| ALPHA | THETA | Mod \|0⟩ | Mod \|1⟩ | Prob (0) | Prob (1) | ProbTotal |
|--------|--------|---------|---------|----------|----------|-----------|
| 0.0000 | 1.5708 | 1.00000 | 0.00000 | 1.00000 | 0.00000 | 1.00 |
| 0.1745 | 1.4835 | 0.99619 | 0.08716 | 0.99239 | 0.00668 | 1.00 |
| 0.3491 | 1.3963 | 0.98481 | 0.17365 | 0.96985 | 0.03015 | 1.00 |
| 0.5236 | 1.3090 | 0.96593 | 0.25882 | 0.93302 | 0.06699 | 1.00 |
| 0.6981 | 1.2217 | 0.93969 | 0.34202 | 0.88301 | 0.11698 | 1.00 |
| 0.8727 | 1.1345 | 0.90631 | 0.42262 | 0.82140 | 0.17861 | 1.00 |
| 1.0472 | 1.0472 | 0.86603 | 0.50000 | 0.75001 | 0.25000 | 1.00 |
| 1.2217 | 0.9599 | 0.81915 | 0.57358 | 0.67101 | 0.32899 | 1.00 |
| 1.3963 | 0.8727 | 0.76604 | 0.64279 | 0.58682 | 0.41318 | 1.00 |
| 1.5708 | 0.7854 | 0.70711 | 0.70711 | 0.50000 | 0.50000 | 1.00 |
| 1.7453 | 0.6981 | 0.64279 | 0.76604 | 0.41318 | 0.58682 | 1.00 |
| 1.9199 | 0.6109 | 0.57358 | 0.81915 | 0.32899 | 0.67101 | 1.00 |
| 2.0944 | 0.5236 | 0.50000 | 0.86603 | 0.25000 | 0.75001 | 1.00 |
| 2.2689 | 0.4363 | 0.42262 | 0.90631 | 0.17861 | 0.82140 | 1.00 |
| 2.4435 | 0.3491 | 0.34202 | 0.93969 | 0.11698 | 0.88302 | 1.00 |
| 2.6180 | 0.2618 | 0.25882 | 0.96593 | 0.06699 | 0.93302 | 1.00 |
| 2.7925 | 0.1745 | 0.17365 | 0.98481 | 0.03015 | 0.96985 | 1.00 |
| 2.9671 | 0.0873 | 0.08761 | 0.99619 | 0.00768 | 0.99240 | 1.00 |
| 3.1416 | 0.0000 | 0.00000 | 1.00000 | 0.00000 | 1.00000 | 1.00 |

*Table 5. Results of the verification process carried out.*

Similarly, Table 6 illustrates the relationships found between Subjective Disbelief, Subjective Credibility and the corresponding quantum probabilities associated to the fact under consideration.

| Subjective Disbelief | Subjective Credibility | Subjective Classification | THETA | Ket 0 | Ket 1 | Prob (True) | Prob (False) | Total Probability |
|------|------|------|------|------|------|------|------|------|
| 0 | 100 | True | 1,57080 | 1,000 | 0,000 | 1,000 | 0,000 | 1,000 |
| 10 | 90 | Almost Certainly True | 1,41372 | 0,988 | 0,156 | 0,976 | 0,024 | 1,000 |
| 20 | 80 | Very Likely | 1,25664 | 0,951 | 0,309 | 0,905 | 0,095 | 1,000 |
| 30 | 70 | Likely | 1,09956 | 0,891 | 0,454 | 0,794 | 0,206 | 1,000 |
| 40 | 60 | Somewhat Likely | 0,94248 | 0,809 | 0,588 | 0,655 | 0,345 | 1,000 |
| 50 | 50 | Unknown | 0,78540 | 0,707 | 0,707 | 0,500 | 0,500 | 1,000 |
| 60 | 40 | Somewhat Unlikely | 0,62832 | 0,588 | 0,809 | 0,345 | 0,655 | 1,000 |
| 70 | 30 | Unlikely | 0,47124 | 0,454 | 0,891 | 0,206 | 0,794 | 1,000 |
| 80 | 20 | Very Unlikely | 0,31416 | 0,309 | 0,951 | 0,095 | 0,905 | 1,000 |
| 90 | 10 | Almost Certainly False | 0,15708 | 0,156 | 0,988 | 0,024 | 0,976 | 1,000 |
| 100 | 0 | False | 0,00000 | 0,000 | 1,000 | 0,000 | 1,000 | 1,000 |

*Table 6. Subjective values and associated quantum probabilities.*

From the previous verification results it can be deduced that, for any rule-based system, the degree of disbelief associated to a given fact can be managed with the following quantum model:

$$|\psi(\theta)\rangle = \left|\psi\left(\frac{\pi-\alpha}{2}\right)\right\rangle = \left|\psi\left(\frac{\pi}{2}\right)\left(1-\frac{\delta}{100}\right)\right\rangle = \sin(\theta)\ |0\rangle + \cos(\theta)\ |1\rangle$$

Where $\alpha \in [0, \pi]$, $\delta \in [0, 100]$ and, in addition, the results can be normalized between 0 and 1. This circumstance could open the door to the quantum processing of uncertain information.

### 7. Experiments and Results

With the same philosophy of the performed verification, and whose results are illustrated in Table 5, a series of experiments were performed for different DELTA values applying the model described in the previous section. The objective of this experimentation is to identify the state function associated to the subjective disbelief, DELTA, which is - from a quantitative point of view - absolutely independent of the rule under consideration and it only depends on the credibility associated with the facts of the premises of the rule. Table 7 illustrates the results of the experimentation.

| DELTA | Prob (True) | Prob (False) | Total Probability | Amplitude (Ket 0) | Amplitude (Ket 1) |
|-------|-------------|--------------|-------------------|-------------------|-------------------|
| 0 | 1,00000 | 0,00000 | 1,00000 | 1,000 | 0,000 |
| 10 | 0,97504 | 0,02496 | 1,00000 | 0,987 | 0,158 |
| 20 | 0,90394 | 0,09606 | 1,00000 | 0,951 | 0,310 |
| 30 | 0,79289 | 0,20711 | 1,00000 | 0,890 | 0,455 |
| 40 | 0,65335 | 0,34665 | 1,00000 | 0,808 | 0,589 |
| 50 | 0,49899 | 0,50101 | 1,00000 | 0,706 | 0,708 |
| 60 | 0,34623 | 0,65377 | 1,00000 | 0,588 | 0,809 |
| 70 | 0,20523 | 0,79477 | 1,00000 | 0,453 | 0,891 |
| 80 | 0,09506 | 0,90494 | 1,00000 | 0,308 | 0,951 |
| 90 | 0,02496 | 0,97504 | 1,00000 | 0,158 | 0,987 |
| 100 | 0,00000 | 1,00000 | 1,00000 | 0,000 | 1,000 |

*Table 7. Experimental results*

With these results, the state functions that we obtain are the following:

- $|\psi_{\delta=0}\rangle = |0\rangle$
- $|\psi_{\delta=10}\rangle = 0.987|0\rangle + 0.158|1\rangle$
- $|\psi_{\delta=20}\rangle = 0.951|0\rangle + 0.310|1\rangle$
- $|\psi_{\delta=30}\rangle = 0.890|0\rangle + 0.455|1\rangle$
- $|\psi_{\delta=40}\rangle = 0.808|0\rangle + 0.589|1\rangle$
- $|\psi_{\delta=50}\rangle = 0.706|0\rangle + 0.708|1\rangle$

- $|\psi_{\delta=60}\rangle = 0.588|0\rangle + 0.809|1\rangle$
- $|\psi_{\delta=70}\rangle = 0.453|0\rangle + 0.891|1\rangle$
- $|\psi_{\delta=80}\rangle = 0.308|0\rangle + 0.951|1\rangle$
- $|\psi_{\delta=90}\rangle = 0.158|0\rangle + 0.987|1\rangle$
- $|\psi_{\delta=100}\rangle = |1\rangle$

As a next step of the experimentation, we decided to apply the developed model to the example of Figure 2, but taking into account the restrictions imposed by the credibility of the facts. This circumstance involves the design and implementation of a new quantum module, $\psi(\theta)$, which replaces the Hadamard Gates, but also puts the system in a state of coherent superposition. With these considerations, the circuit of Figure 2 is modified as illustrated in Figure 7.

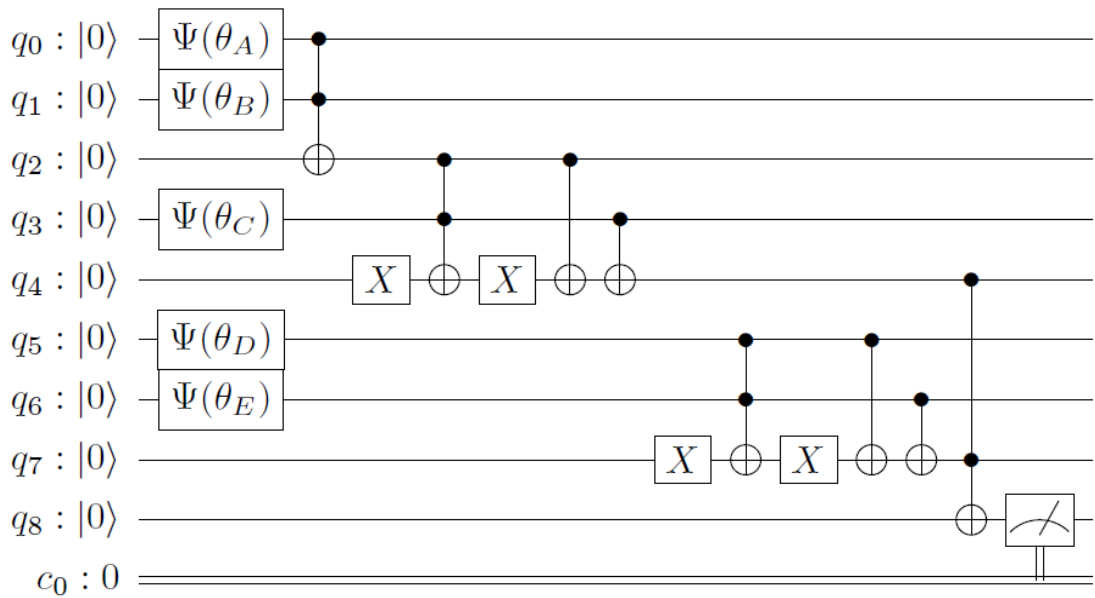

*Figure 7. Quantum circuit analogous to that of figure 2, but in which the gates H have been replaced by gates $\psi(\theta)$.*

To check the consistency of the model in this inferential circuit, we consider the following $\delta$ arbitrary for the facts A, B, C, D and E of the example:

- $\delta_A = 50$
- $\delta_B = 50$
- $\delta_C = 50$
- $\delta_D = 50$
- $\delta_E = 50$

The results obtained after the measurement were the following:

- Probability of a true conclusion = 0.53205
- Probability of a false conclusion = 0.46795

These results are exactly the same as those we would obtain if we solved in a traditional way the circuit of Figure 1 (from which the other two quantum circuits derive). Obviously, the Total Probability is 1,000 [Moret-Bonillo, 20018]

Finally, we carried out a new experiment, this time with different values of the DELTA parameters associated with the different facts of the rules of our example. The results obtained, illustrated in table 8, show that the approximation followed and the proposed model produce coherent results.

| DELTA A | DELTA B | DELTA C | DELTA D | DELTA E | Prob (True) | Prob (False) | Prob (Total) |
|---------|---------|---------|---------|---------|-------------|--------------|--------------|
| 0 | 0 | 20 | 0 | 0 | 1,000 | 0,000 | 1,00 |
| 20 | 60 | 0 | 0 | 20 | 0,994 | 0,006 | 1,00 |
| 40 | 80 | 0 | 20 | 20 | 0,945 | 0,055 | 1,00 |
| 60 | 100 | 20 | 0 | 0 | 1,000 | 0,000 | 1,00 |
| 80 | 80 | 20 | 0 | 20 | 0,920 | 0,080 | 1,00 |
| 100 | 60 | 20 | 20 | 20 | 0,874 | 0,127 | 1,00 |
| 0 | 40 | 0 | 0 | 0 | 1,000 | 0,000 | 1,00 |
| 20 | 20 | 0 | 0 | 20 | 1,000 | 0,000 | 1,00 |
| 40 | 0 | 0 | 20 | 0 | 1,000 | 0,000 | 1,00 |
| 60 | 20 | 0 | 20 | 20 | 0,989 | 0,011 | 1,00 |
| 80 | 40 | 20 | 0 | 0 | 1,000 | 0,000 | 1,00 |
| 100 | 60 | 20 | 0 | 20 | 0,936 | 0,064 | 1,00 |
| 0 | 80 | 20 | 20 | 0 | 0,991 | 0,009 | 1,00 |
| 20 | 100 | 20 | 20 | 20 | 0,967 | 0,033 | 1,00 |
| 40 | 80 | 0 | 0 | 0 | 1,000 | 0,000 | 1,00 |
| 60 | 60 | 0 | 0 | 20 | 0,957 | 0,043 | 1,00 |
| 80 | 40 | 0 | 20 | 0 | 0,968 | 0,032 | 1,00 |
| 100 | 20 | 0 | 20 | 20 | 0,985 | 0,015 | 1,00 |
| 0 | 0 | 20 | 0 | 0 | 1,000 | 0,000 | 1,00 |
| 20 | 100 | 20 | 0 | 20 | 0,984 | 0,016 | 1,00 |
| 40 | 80 | 20 | 20 | 80 | 0,657 | 0,343 | 1,00 |
| 60 | 60 | 20 | 60 | 20 | 0,671 | 0,329 | 1,00 |
| 0 | 0 | 0 | 0 | 0 | 1,000 | 0,000 | 1,00 |
| 20 | 20 | 20 | 20 | 20 | 0,981 | 0,019 | 1,00 |
| 40 | 40 | 40 | 40 | 0 | 0,854 | 0,146 | 1,00 |
| 60 | 60 | 60 | 0 | 20 | 0,927 | 0,073 | 1,00 |
| 80 | 80 | 20 | 20 | 0 | 0,914 | 0,086 | 1,00 |
| 100 | 100 | 100 | 100 | 100 | 0,000 | 1,000 | 1,00 |

*Table 8. Experimental results obtained with different DELTA values associated to the facts of the example.*

**8. Discussion and Conclusions**

It is an obvious fact that uncertainty, in the most general sense possible and from any point of view, is a problem of the first magnitude - still unresolved - in the field of artificial intelligence and, more specifically, in Rule-Based Systems. Regardless of the completeness of our system, the inherent subjectivity of the uncertainty associated with the information we use when trying to solve a real case, has involved the development of a multitude of approaches and models to try to solve this problem. The mathematical orientation of the different approaches varies depending on the model in question. Therefore, faced with the same problem, different models produce different results. In this context we can mention, among others:

- Categorical approaches such as the so-called Differential Interpretation [Ledley, 1959]
- Probabilistic approaches such as Bayesian Networks [Pearl, 1986]
- Quasi-statistical approaches such as the Certainty Factors Method [Shortliffe, 1975] or Evidential Theory [Shafer, 1976]
- Fuzzy methods such as Fuzzy Logic [Zadeh, 1965]

In relation to this problem, we cannot forget the potential of emerging theories and applications, among which Quantum Computing stands out and it is intrinsically probabilistic. The question is, therefore, how can we model the subjective uncertainty of rule-based systems and achieve coherent results using the resources of quantum computing? In short, it is about establishing synergies between artificial intelligence and quantum computing to solve the problem of uncertainty.

In this work we have focused the question from the perspective of the Theory of Quantum Circuits. For this we built the quantum operators equivalent to the classic AND and OR operators. We also developed a classic inferential circuit and built the equivalent quantum circuit. To model the uncertainty we propose the use of Z displacements in the Bloch sphere, and to model the subjectivity we introduce the parameter DELTA ($\delta$). We then build a quantum gate, which is of a universal nature, in order to represent any situation of uncertainty and express it in a probabilistic manner. The proposed approach is widely evaluated and verified from different points of view. The results obtained after this exhaustive validation process allow us to conclude that Quantum Computation is an effective and efficient method to solve uncertainty problems in Artificial Intelligence.

**Acknowledgements:** This work has received financial support from the Xunta de Galicia (Centro singular de investigación de Galicia accreditation 2016-2019) and the European Union (European Regional Development Fund - ERDF). Financial support from the Xunta de Galicia (Centro singular de investigación de Galicia accreditation 2016-2019) and the European Union (European Regional Development Fund - ERDF), is gratefully acknowledged.